\documentclass[runningheads]{llncs}
\usepackage[T1]{fontenc}
\usepackage{graphicx}
\usepackage{hyperref}
\usepackage{color}

\usepackage{todonotes}
\setuptodonotes{inline}

\usepackage{subcaption}
\usepackage{pifont}

\begin{document}
\title{GreenStableYolo: Optimizing Inference Time and Image Quality of Text-to-Image Generation}
\titlerunning{GreenStableYolo}
%
\author{Jingzhi Gong\inst{1}\orcidID{0000-0003-4551-0701} \and
Sisi Li\inst{2}\orcidID{0000-0001-5376-5144} \and
Giordano d'Aloisio\inst{3}\orcidID{0000-0001-7388-890X}\and
Zishuo Ding\inst{4}\orcidID{0000-0002-0803-5609}\and
Yulong Ye\inst{5}\orcidID{0000-0002-2478-778X}\and
William B. Langdon\inst{6}\orcidID{0000-0002-6388-4160}\and
Federica Sarro\inst{6}\orcidID{0000-0002-9146-442X}}
\authorrunning{J. Gong et al.}

\institute{Loughborough University
\email{j.gong@lboro.ac.uk}\and
Beijing University of Posts and Telecommunications
\email{sisili@bupt.edu.cn} \and
Università degli Studi dell'Aquila
\email{giordano.daloisio@graduate.univaq.it} \and
University of Waterloo
\email{zishuo.ding@uwaterloo.ca} \and
University of Birmingham
\email{yxy382@student.bham.ac.uk} \and
University College London
\email{\{w.langdon, f.sarro\}@ucl.ac.uk} }

\maketitle              

\vspace{-0.2cm}
\begin{abstract}
Tuning the parameters and prompts for improving AI-based text-to-image generation has remained a substantial yet unaddressed challenge. Hence we introduce GreenStableYolo, which improves the parameters and prompts for Stable Diffusion to both reduce GPU inference time and increase image generation quality using NSGA-II and Yolo. Our experiments show that despite a relatively slight trade-off (18\%) in image quality compared to StableYolo (which only considers image quality), GreenStableYolo achieves a substantial reduction in inference time (266\% less) and a 526\% higher hypervolume, thereby advancing the state-of-the-art for text-to-image generation.
\end{abstract}

\section{Introduction}
In recent years Generative Artificial Intelligence (GenAI) has emerged as a powerful approach encompassing various techniques that enable machines to generate new content, such as text~\cite{DBLP:journals/corr/abs-2105-10311}, images~\cite{DBLP:conf/cvpr/RombachBLEO22}, and videos~\cite{DBLP:journals/corr/abs-2401-01256}. Particularly, image generation and text-to-image synthesis have garnered significant attention due to their potential in bridging the gap between textual descriptions and visual representations~\cite{DBLP:conf/icml/ReedAYLSL16}. It enables systems to understand and interpret human language and automatically translate it to a visually meaningful way, facilitating tasks such as generating accompanying images for books~\cite{BRUNS2024103790}, generating product images for advertising~\cite{DBLP:conf/icde/YuYJZZL21}, and inspiring artists to create new forms of art~\cite{DBLP:conf/iui/KoPJJKS23}.

However, achieving optimal performance in image generation tasks involves fine-tuning various aspects of a GenAI model, such as the number of inference steps, positive and negative prompts~\cite{stable_diffusion_docs,DBLP:conf/ssbse/BergerDDEKMMS23}. These parameters play a crucial role in determining the quality and efficiency of the generated images and tuning these parameters is essential to unlock the full potential of image generation models~\cite{DBLP:conf/ssbse/BergerDDEKMMS23,DBLP:conf/icdsc/MaglianiSC019}.  At the same time, GenAI model are energy demanding and largely contribute to the increased CO2 emissions\cite{10.1145/3510003.3510221,SarroRE}.



Berger et al.~\cite{DBLP:conf/ssbse/BergerDDEKMMS23} proposed a search-based approach, dubbed StableYolo, to optimize the image quality of Stable Diffusion by assessing image quality using Yolo~\cite{DBLP:conf/cvpr/RedmonDGF16}. However, their approach do not take into account the aspect of inference time, which is a cornerstone for both ensuring user experience and minimizing the energy consumption of GenAI models. Especially in real-world scenarios, where responsiveness and energy efficiency are vital, addressing this aspect is vital for the widespread adoption of such models~\cite{DBLP:journals/corr/abs-2303-04226,DBLP:journals/corr/abs-2302-14017,SarroRE,sarroICPE}.

To address this gap, we present GreenStableYolo, a novel approach that addresses the challenge of optimizing the trade-off between inference time and image quality using a search-based multi-objective optimization method, namely Non-dominated Sorting Genetic Algorithm (NSGA-II)~\cite{DBLP:journals/tec/DebAPM02}. We provide initial empirical evidence that by using GreenStableYolo Stable Diffusion achieves a satisfactory equilibrium between inference time and image quality, making it suitable for real-world applications where both factors play a crucial role. 

In a nutshell, the key contributions of this work include:

\begin{itemize}
    \item The development of a novel system that seeks for an optimal trade-off between inference time and image quality by optimizing the prompts and parameters for Stable Diffusion, dubbed GreenStableYolo;
    \item Empirical evidence on the effectiveness of GreenStableYolo in achieving significantly less inference time and higher hypervolume compared to StableYolo, thereby advancing the state-of-the-art multi-objective optimization for text-to-image generation;
    \item A comprehensive analysis to understand the influence of different parameters and prompts on both inference time and image quality in Stable Diffusion.
\end{itemize}

\section{Related Work}
To improve image generation quality, Berger et al.~\cite{DBLP:conf/ssbse/BergerDDEKMMS23} were the first to propose the use of a Genetic Algorithm (GA) able to simultaneously tune the prompt and parameters of Stable Diffusion. Magliani et al.~\cite{DBLP:conf/icdsc/MaglianiSC019} use GA to find the best diffusion parameters for automated image retrieval from a dataset. 
While some research~\cite{DBLP:journals/corr/abs-2303-04226,DBLP:journals/corr/abs-2302-14017} has been carried out to optimize inference time, from hardware design to model architecture, there has been limited work focusing on optimizing the prompts and parameters. Our work builds upon previous work by considering both inference time and image quality as optimization objectives.

%

\section{Methodology}
To mitigate the aforementioned challenge, we propose GreenStableYolo, a novel multi-objective search-based approach that, given a text prompt for image generation, searches for the optimal parameters that can strike a balance between:
(1) \textit{Inference time}, which is measured by the GPU time taken for the execution of the StableDiffusionPipeline; and (2) \textit{Image quality}, which is determined by performing object recognition with Yolo, then selecting objects that match the input prompt, and computing their average probabilities~\cite{DBLP:conf/ssbse/BergerDDEKMMS23}.

\vspace{0.3cm}
\noindent \textbf{NSGA-II Optimization Algorithm}
To simultaneously enhance image quality and reduce inference time, we leverage NSGA-II, a well-known and efficient multi-objective evolutionary algorithm~\cite{DBLP:journals/tits/JiHY23,DBLP:journals/access/RashidMTBDHDA24}. Specifically, NSGA-II works as follows: 
(1) Initialize a population with $N$ individuals; 
(2) Perform crossover and mutation operations, generating an offspring population denoted as $P_{o}$;
(3) Reassemble the parent population $P_{t-1}$ and $P_{o}$ into a temporary population with the size of $2N$, and formulate individuals into $i$ non-inferior frontier through fast non-dominating sorting;
(4) Select $N$ individuals from the temporary population to form the next population for the $t^{th}$ iteration, denoted as $P_{t}$.
(5) Repeat steps (2)-(4) until the termination condition is met; and
(6) The algorithm ends up and returns the current Pareto-Optimal set.

\vspace{0.3cm}
\noindent \textbf{Selected Parameters }
To make a straightforward comparison with StableYolo, we adopt the same settings as used by Berger et al.~\cite{DBLP:conf/ssbse/BergerDDEKMMS23}. Specifically, the following parameters and prompts are tuned and searched with NSGA-II:
(1) \textit{Inference steps} (1 to 100): the AI's image generation iterations;
(2)~\textit{Guidance scale} (1 to 20): the impact of the prompt on image generation;
(3) \textit{Guidance rescale} (0 to 1): rescales the guidance factor to prevent over-fitting;
(4) \textit{Seed} (1 to 512): randomization seed;
(5) \textit{Positive prompt}: used to describe images and improve their details, e.g., ``photograph'', ``color'', and ``ultra real''~\cite{stable_diffusion_prompts}; and
(6) \textit{Negative prompt}: avoided description during image generation, e.g., ``sketch'', ``cropped'', and ``low quality''~\cite{stable_diffusion_prompts}.


\section{Evaluation}

To evaluate our proposal, we address the following research questions (RQs):

\noindent\ding{228} \textbf{RQ1: }To what extent can GreenStableYolo improve image quality and inference time compared with StableYolo?

\noindent\ding{228} \textbf{RQ2: }How do parameters/prompts of Stable Diffusion influence the inference time for image generation?

\noindent\ding{228} \textbf{RQ3: }How do parameters/prompts of Stable Diffusion influence the quality of the generated images?

\vspace{0.3cm}
\noindent \textbf{Experimental Setup }
To ensure a fair evaluation of the optimization effectiveness, we employed the same hyperparameter setup as StableYolo for NSGA-II, e.g., the population size was set to 25, the number of generations was set to 50, and both the mutation rate and crossover rate were set to 0.2. {We selected the weights of 0.001 for image quality and -1000 for inference time based on empirical investigation of different weight combinations.} In addition, we used Stable Diffusion version~v2 and Yolo version~v8. 
To assess variability, we evaluated each model 15 times using different random seeds, focusing solely on the prompt ``two people and a bus'' due to time constraints. Any future studies can explore additional prompts. All experiments were conducted on a virtual machine hosted on Google Colaboratory, with an NVIDIA Tesla T4 GPU with 16 GB of RAM.



\begin{figure}[t!]
    \centering
    \begin{subfigure}{.32\textwidth}
        \centering
        \includegraphics[width=\textwidth]{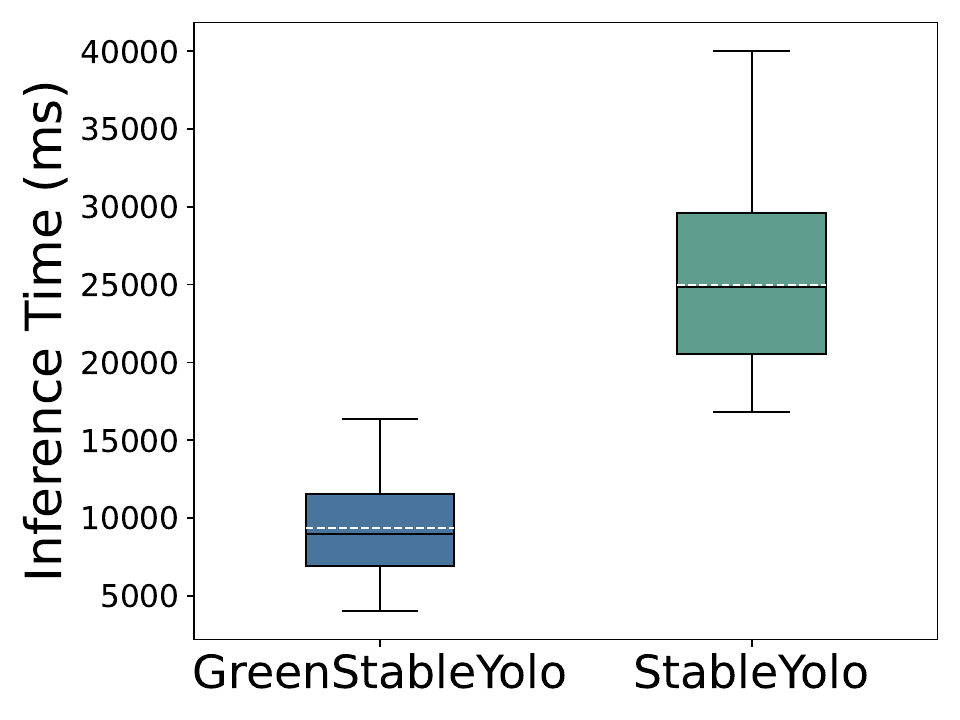}
        \caption{Inference time}
        \label{fig:rq1-a}
    \end{subfigure}
    \begin{subfigure}{.32\textwidth}
        \centering
        \includegraphics[width=\textwidth]{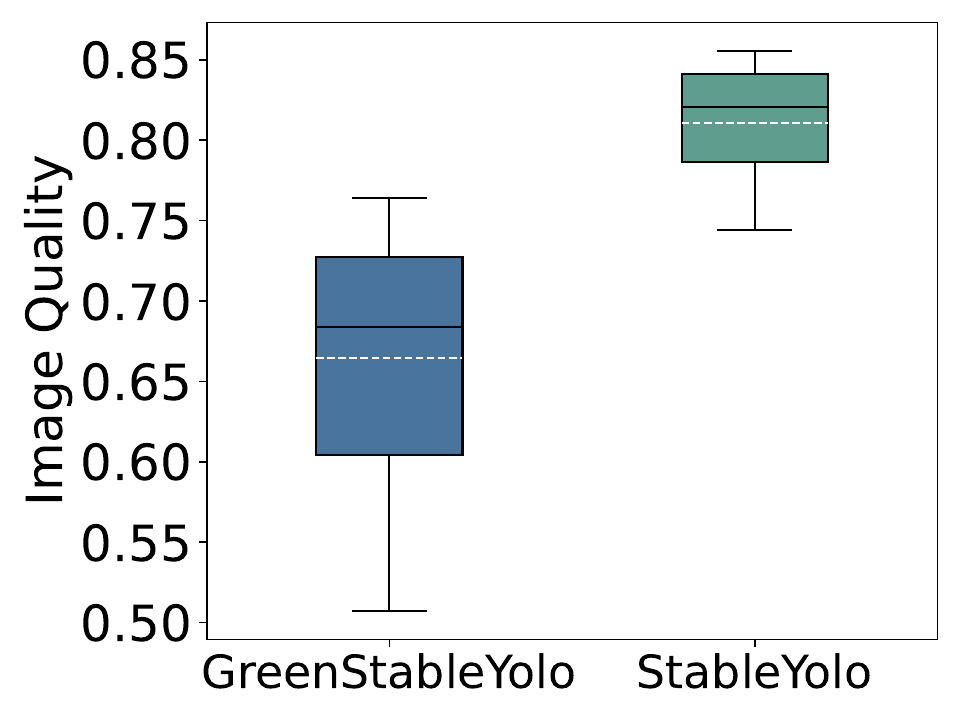}
        \caption{Yolo quality measure}
        \label{fig:rq1-b}
    \end{subfigure}
    \begin{subfigure}{.32\textwidth}
        \centering
        \includegraphics[width=\textwidth]{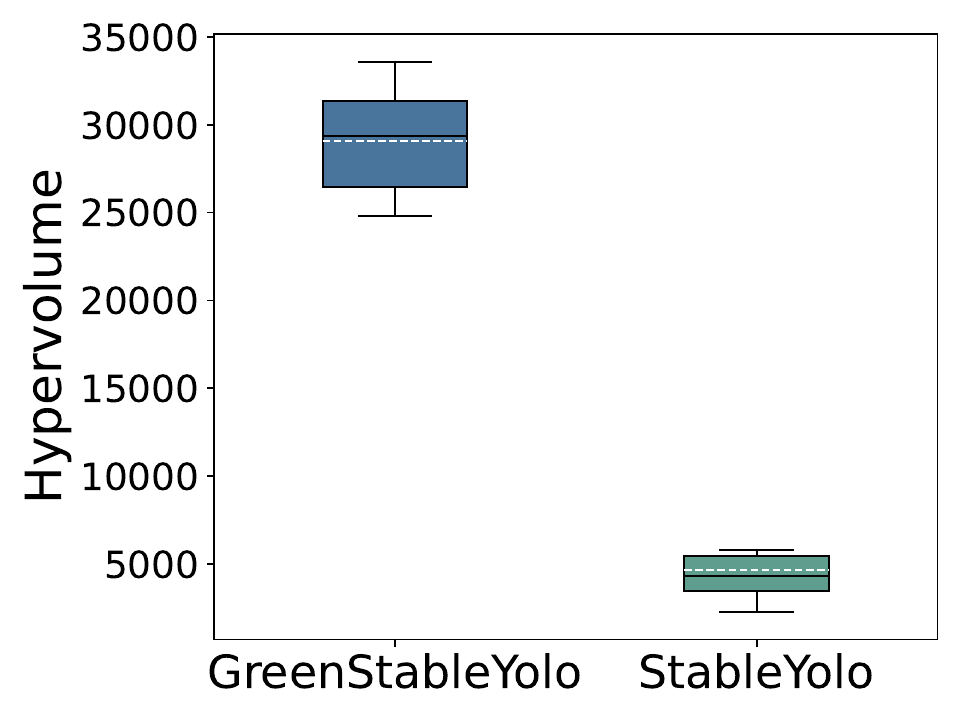}
        \caption{Hypervolume}
        \label{fig:rq1-c}
    \end{subfigure}
    \vspace{-2mm}
    \caption{Comparison of GreenStableYolo and StableYolo on 15 independent runs}
    \label{metricbar}
\end{figure}

\vspace{0.3cm}
\noindent \textbf{RQ1 Results }
Figure~\ref{metricbar} presents the performance comparisons between GreenStableYolo and StableYolo. Specifically, Figure~\ref{fig:rq1-a} reveals that GreenStableYolo achieves an average inference time of 
9.4~seconds 
with an interquartile range~(IQR) of 
4.7~seconds. 
Conversely, StableYolo exhibits an average inference time of 25.0~seconds, 
which is 1.66 times higher than GreenStableYolo, with an IQR of 9.1~seconds. 
That is, GreenStableYolo generates images much faster. 

This improvement in inference time comes at a slight cost to image quality. 
As illustrated in Figure~\ref{fig:rq1-b}, GreenStableYolo experiences approximately an average degradation of 0.18 points in image quality.
We also compute the hypervolume~\cite{guerreiro2021hypervolume} for both models for a more comprehensive comparison\footnote{Hypervolume is a fundamental metric used in multi-objective optimization problems that indicates the dominance of a solution in the objective space.}. Figure~\ref{fig:rq1-c} presents the hypervolume values with the reference point set as (1, 50000), where GreenStableYolo achieves an average hypervolume of 29074.11, surpassing StableYolo's score of 4642.17 by 5.26 times. \emph{\textbf{This substantial difference demonstrates the clear dominance of GreenStableYolo over StableYolo in this two-objective optimization problem for text-to-image generation.}}

\vspace{0.3cm}
\noindent \textbf{RQs2-3 Results }
To investigate RQs2--3, we followed previous work~\cite{10.1145/3377811.3380351} and built two Random Forest regression models using \texttt{scikit-learn}. The features of these models include the number of iteration steps, guidance scale, guidance rescale, positive prompts, and negative prompts (excluding the random seed). The target variables are inference time and image quality score, respectively. We use the \texttt{RandomizedSearchCV} function from \texttt{scikit-learn} to find the optimal hyperparameters during model training. The \texttt{feature\_importances\_} function is then used to compute the importance of each parameter and prompt based on the Mean Decrease Impurity (MDI), a.k.a. as Gini importance. To ensure reliability, we repeat this process 10 times.

\begin{figure}[t]
    \centering
        \begin{subfigure}{.45\textwidth}
        \centering
        \includegraphics[width=\textwidth]{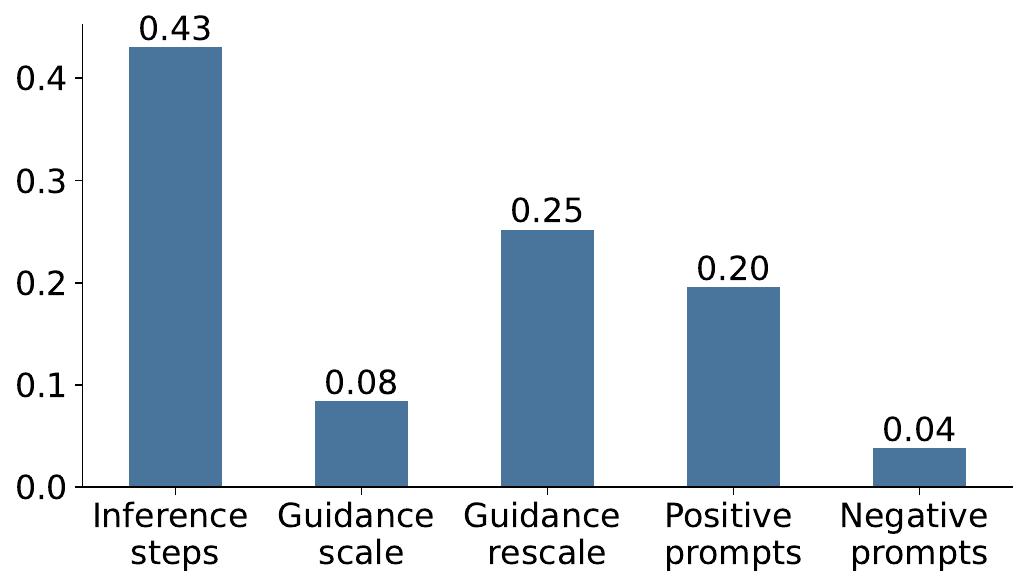}
        \caption{RQ2: MDI w.r.t. inference time}
        \label{fig:rq2}
    \end{subfigure}
    \begin{subfigure}{.45\textwidth}
        \centering
        \includegraphics[width=\textwidth]{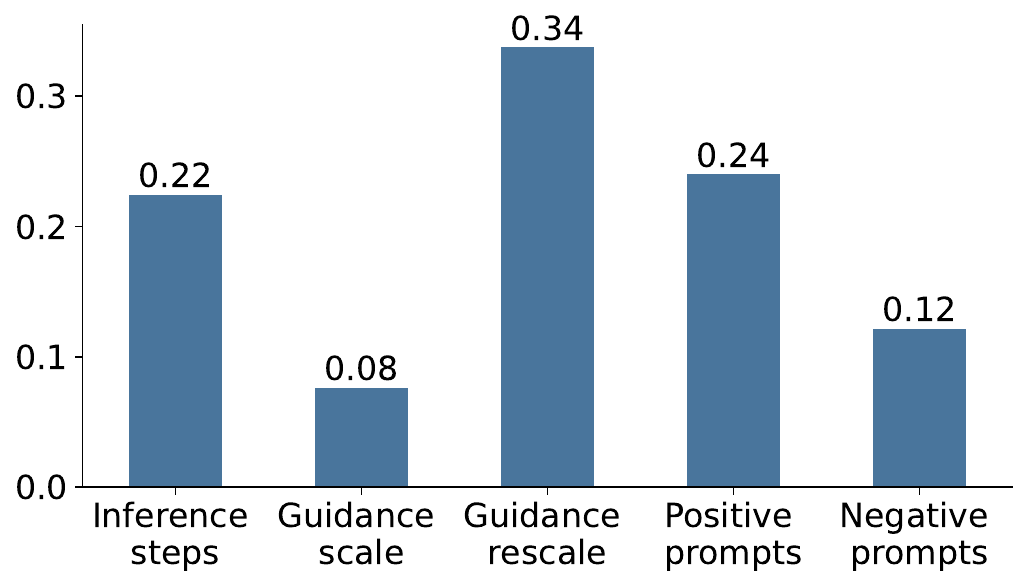}
        \caption{RQ3: MDI w.r.t. image quality}
        \label{fig:rq3}
    \end{subfigure}

    \caption{Parameters/prompts importance based on the mean decrease in impurity}
    \label{fig:rq2-3}
\end{figure}

Figures~\ref{fig:rq2} and \ref{fig:rq3} present the calculated importance of parameters and prompts based on the mean decrease in impurity, with respect to inference time and image quality scores, respectively. As shown in Figure \ref{fig:rq2}, the number of \textit{inference steps} emerges as a significant factor affecting inference time. This is expected, as more steps involve more computations, thereby resulting in higher inference time. Meanwhile, for image quality (Figure \ref{fig:rq3}), parameters like \textit{guidance rescale} and \textit{positive prompts} play a relatively more critical role.

These findings confirm the value of our work: \emph{\textbf{increased computational resources do not necessarily translate to better image quality; instead, appropriate model parameter settings are more crucial}}. This highlights the importance of identifying optimal parameter combinations during model inference to balance computational efficiency and output quality.

\vspace{0.3cm}
\noindent \textbf{Threats to Validity} The limited exploration of prompts, the randomness in the optimization process, and the specific configuration for NSGA-II may introduce {internal threats}. Besides, {external threats} may include the choice of the GenAI model, the noise when measuring the inference time, and the evaluation of image quality based on object recognition using Yolo. 

\section{Conclusion}
In GenAI text-to-image, achieving images of high-quality is often not the only important aspect to consider, as inference time, which directly impacts user experience and energy consumption, also plays a critical role. 
In this work we introduced GreenStableYolo, the first approach leveraging NSGA-II to strike an optimal trade-off between these two objectives for Stable Diffusion. Experimental comparisons with StableYolo demonstrate that GreenStableYolo achieves significantly reduced inference time while maintaining a relatively high image quality. Future research can expand upon our evaluation by incorporating alternative initial prompts, optimizing different performance metrics such as energy consumption, and broadening to other GenAI systems such as DALL-E, ImageFX, or Midjourney.

\vspace{0.3cm}
\noindent \textbf{Availability }
Repository available at \url{https://github.com/gjz78910/GreenStableYolo}.

\vspace{0.3cm}
\noindent \textbf{Acknowledgments }
Work supported by the ERC grant no. 741278.

%
\bibliographystyle{splncs04}
\bibliography{references}

\end{document}